\documentclass[lettersize,journal]{IEEEtran} 

\usepackage{amsmath,amsfonts}
\usepackage{algorithmic}
\usepackage{array}
\usepackage{textcomp}
\usepackage{stfloats}
\usepackage{url}
\usepackage{verbatim}
\usepackage{graphicx}
\def\BibTeX{{\rm B\kern-.05em{\sc i\kern-.025em b}\kern-.08em
    T\kern-.1667em\lower.7ex\hbox{E}\kern-.125emX}}
\usepackage{balance}
\usepackage{latexsym}
\usepackage{amssymb}
\usepackage{amsmath}
\usepackage{amsthm}
\usepackage{booktabs}
\usepackage{enumitem}
\usepackage{color}
\usepackage{times}
\usepackage{epsfig}
\usepackage{amsmath}
\usepackage{amssymb}
\usepackage{multirow}
\usepackage{booktabs}
\usepackage{colortbl}
\usepackage{bbding}
\usepackage{pifont}
\usepackage{wasysym}
\usepackage{amssymb}
\usepackage{graphicx}
\usepackage{enumitem}
\usepackage[flushleft]{threeparttable}
\usepackage[normalem]{ulem}
\usepackage{setspace}
\usepackage{pifont}
\usepackage{rotating}
\usepackage{longtable,booktabs,threeparttablex}
\usepackage{threeparttable} 
\usepackage{comment}
\usepackage{booktabs}
\usepackage{longtable}
\usepackage{tabularray}
\usepackage{xtab}
\usepackage{cite}
\usepackage[font=footnotesize,labelfont=sf,textfont=sf]{subcaption}
\usepackage{caption}

\begin{document}

\title{A Comparative Study of Continuous Sign Language Recognition Techniques}
\author{Sarah Alyami$^{1,2}$ and Hamzah Luqman$^2$ \\
$^1$Applied College, Imam Abdulrahman Bin Faisal University \\
$^2$Information and Computer Science Department, King Fahd University of Petroleum and Minerals\\ 
SDAIA-KFUPM Joint Research Center for Artificial Intelligence, KFUPM, Saudi Arabia \\
 {snalyami@iau.edu.sa, hluqman@kfupm.edu.sa}}


\maketitle

\begin{abstract}
Continuous Sign Language Recognition (CSLR) focuses on the interpretation of a sequence of sign language gestures performed continually without pauses.   In this study, we conduct an empirical evaluation of recent deep learning CSLR techniques and assess their performance across various datasets and sign languages. The models selected for analysis implement a range of approaches for extracting meaningful features and employ distinct training strategies. To determine their efficacy in modeling different sign languages, these models were evaluated using multiple datasets, specifically RWTH-PHOENIX-Weather-2014, ArabSign, and GrSL, each representing a unique sign language. The performance of the models was further tested with unseen signers and sentences. The conducted experiments establish new benchmarks on the selected datasets and provide valuable insights into the robustness and generalization of the evaluated techniques under challenging scenarios. 
\end{abstract}


\section{Introduction}
\label{sec:intro}
\noindent
Sign language serves as a critical means of communication delivered using visual cues, including hand gestures and facial expressions~\cite{suliman2021arabic}. Sign language recognition (SLR) involves interpreting signs in video sequences and converting them into their corresponding glosses. The process includes capturing the movements of the signer's hands and body which are usually integrated with facial expressions ~\cite{Wadhawan2021}. SLR aims to facilitate communication between deaf individuals and the broader community by converting sign language into a form that can be understood by those who do not know sign language, thus breaking down communication barriers and promoting inclusively. 

SLR can be mainly classified into isolated SLR and continuous SLR (CSLR). Isolated SLR aims to recognize a single sign from a video clip, whereas continuous SLR (CSLR) interprets a series of signs and produces their corresponding labels i.e. glosses ~\cite{ALYAMI2024103774}. CSLR holds greater societal benefit for real-world usage since it involves recognizing the sequence of signs as they naturally flow together in conversation.  

A significant challenge in CSLR is understanding the sign gestures with their context. Continuous sign language sentences usually involve finger-spelled, static, and dynamic signs. Most finger-spelled signs are static, where no motion is used to express these signs. In contrast, dynamic signs depend on hands and body motion with a variety of non-manual signals, such as facial expressions, performed simultaneously with the sign gestures. There is also often variability in performing signs by different signers which adds more complexity to the CSLR systems.

CSLR is categorized as a weakly supervised learning task due to the imprecise alignment between video frames and their corresponding annotations. The consecutive signs in the sign language sentence are performed continually without clear boundaries. 
Therefore, the exact start and end positions of each sign gesture in the sentence are not clearly defined, and hence need to be learned by the CSLR system ~\cite{Aloysius2020}. Learning signs' boundaries is a challenging task given the lack of boundaries between sentence signs.

\begin{figure} [!ht]
    \centering
    \includegraphics[width=\linewidth]{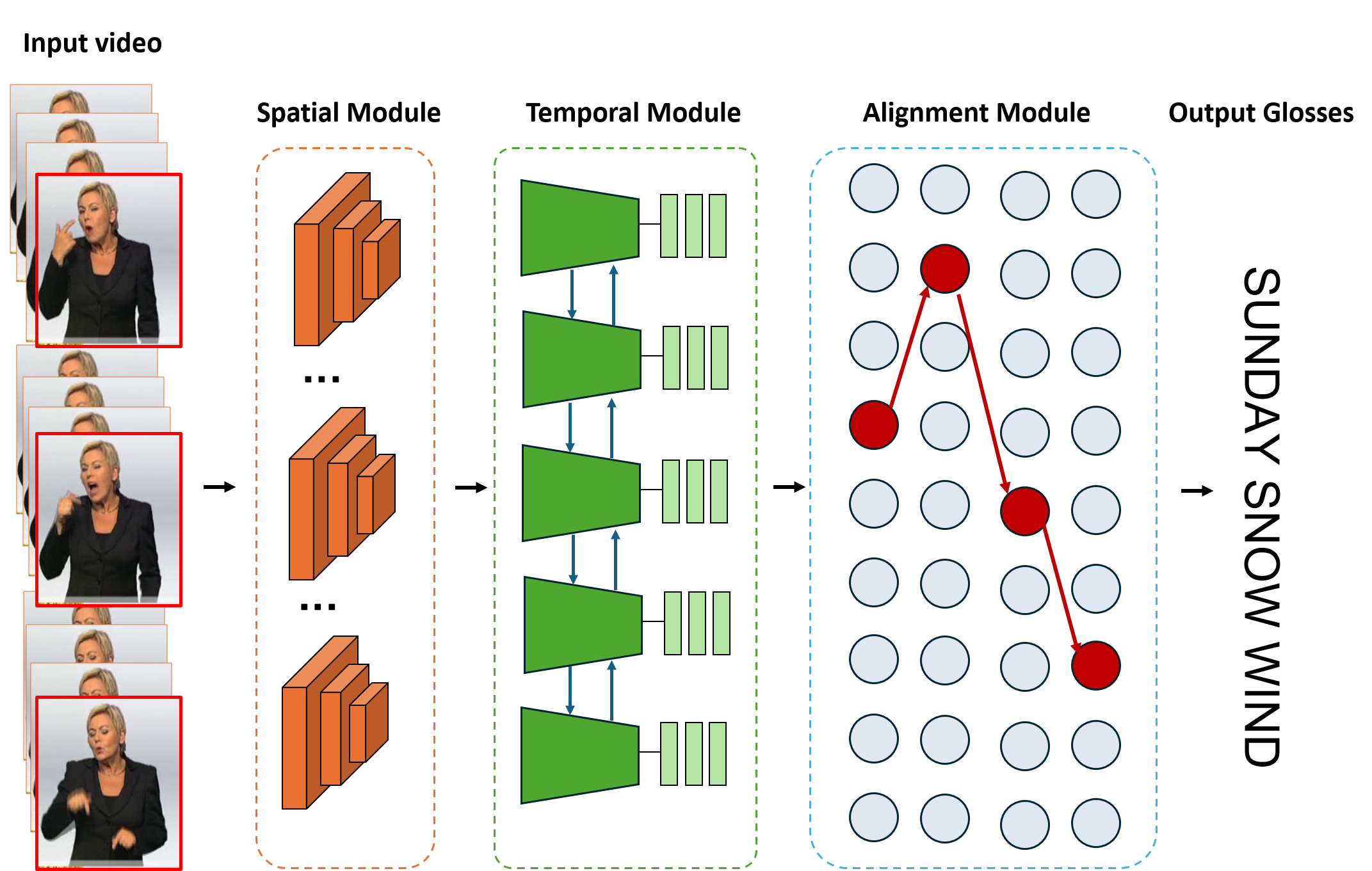}
    \caption{The general framework of CSLR. }
    \label{fig:general_framework}
    
\end{figure}

CSLR systems consist of four main stages: preprocessing the input video stream, extracting the spatial and temporal features from the frame sequence, and learning the proper alignment between frames and glosses, as shown in figure \ref{fig:general_framework}. The preprocessing stage typically involves resizing and normalization of the video frames. Some CSLR systems depend on the skeleton or pose information as an input to the following stages, therefore, they extract the pose information from each video's frame at the preprocessing stage. However, this stage may be skipped by some CSLR systems that utilize sensor-based systems for signs capturing \cite{El-Alfy2022}.  
The spatial feature extractor captures feature representations from sign's frames using spatial information learning techniques, such as 2D Convolutional Neural Networks (CNN), 3D CNNs, Graph Convolutional Networks (GCN) or Vision Transformers (ViT). This step is usually followed by learning the temporal information of the sign gestures using temporal learning techniques, such as Recurrent Neural Networks (RNN) and Temporal Convolutional Networks (TConv). Learning the frame-gloss alignments is performed usually using Hidden Markov Models (HMMs), Connectionist Temporal Classification (CTC), or Dynamic Time Warping (DTW) techniques. Among these, CTC has generally demonstrated more robust performance, and a significant portion of CSLR studies have leveraged CTC as the main training criterion for sequence alignment learning ~\cite{Aloysius2020}. 

The CSLR systems can be evaluated using three distinct protocols: signer-dependent (Signer-Dep), signer-independent (Signer-Indep), and unseen sentences (Unseen-Sent). In \textit{Signer-Dep} evaluation, the CSLR model is evaluated on sentences performed by the same set of signers involved in training data.  The models are exposed to the specific appearances and signing styles of a group of signers during the training and evaluation. Although a high recognition accuracy can be achieved with this evaluation protocol, models may not generalize well to new signers whose data was not included in the training set, as it may overfit the specific features of the training signers. \textit{Signer-Indep} evaluation, on the other hand, involves training the CSLR models on a group of signers, then evaluating them on a completely different set of signers who were not seen during training. The Signer-Indep evaluation is essential for creating inclusive systems that are accessible and usable by a broader and more diverse population of sign language users. In addition, this evaluation helps to develop real-time CSLR systems as it evaluates the generalization of the CSLR systems to new signers.
The third evaluation protocol is to recognize the sign language sentences not available in the training set. The \textit{Unseen-Sent} evaluation is designed to simulate real-world conditions where a CSLR system must be able to accurately recognize and translate signs or sentences that it has never encountered before. This evaluation is more challenging than other evaluation protocols. However, this evaluation is important to assess the system's ability to recognize sign language sentences not present in the training dataset.

Although there has been a surge in CSLR methods, the majority of the state-of-the-art (SOTA) models are trained and assessed using few benchmark datasets, specifically RWTH-PHOENIX-Weather-2014 (Phoenix2014) dataset ~\cite{Koller2015} for German sign language (GSL) and CSL dataset ~\cite{Huang2018} for Chinese sign language (CSL). While there are additional datasets that represent other sign languages, including Arabic (ArSL) ~\cite{luqman2022arabsign}, Greek (GrSL) ~\cite{Adaloglou2021}, and Russian (RSL) \cite{Mukushev2022}. These resources are underutilized by researchers, and the performance of the SOTA methods on other sign languages is unknown. The hesitancy to leverage other CSLR datasets can be attributed mainly to the shortage of established benchmarks on these datasets, which hinders the ability to perform comparisons with newly developed models. As a result, less popular sign languages are under-studied by the research community. This comparative study aims to remedy this issue by targeting less utilized datasets, mainly ArabSign ~\cite{luqman2022arabsign} for Arabic sign language and continuous GrSL ~\cite{Adaloglou2021}. Moreover, we provide insights on the performance of SOTA models in different evaluation settings including Signer-Dep, Signer-Indep, and Unseen-Sent, which were not reported by other researchers. 

This paper is organized as follows. Section \ref{sec:LR} reviews the recent works of CSLR. An overview of the models involved in this comparative study is presented in Section \ref{sec:CS} and the conducted experiments with analysis and discussion are presented in Section \ref{Sec:Ex}. Section \ref{sec:conc} presents our conclusion.   

\section{Literature Review}
\label{sec:LR}
\noindent
Earlier research in CSLR often relied on hand-crafted features and HMM for aligning recognized sentences with their corresponding glosses~\cite{Kelly2009, Forster2013, Zhang2016}. As deep learning advanced, the adoption of CNNs became prevalent for spatial analysis in these systems   \cite{hu2023adabrowse,jang2023self,zheng2023cvt,guo2023distilling,Xie2023,Hu2023,hu2023self,Hu2023multilingual,Zuo2024,Hu2024}. Several hybrid models were introduced for CSLR by combining CNNs with HMMs ~\cite{Koller2016DeepSign,Koller2016}. A hybrid CNN-HMM model was presented in \cite{Koller2016DeepSign} for CSLR using cropped images of the signer's right hand. HamNoSys annotations were added in a later work to model sign language subunits using  hand and mouth shape annotations  ~\cite{Koller2016}. Despite the power of HMMs in sequence learning, HMMs struggled to grasp the long term temporal dependencies in the sign language videos. To overcome this issue, RNNs and CTC have been widely adopted for sequence learning, facilitating end-to-end training that requires only sentence-level labeling ~\cite{Jiao_2023_ICCV,guo2023distilling,Papastratis2021,Hu2023,hu2022temporal,zheng2023cvt,jang2023self,hu2023adabrowse,HU2024109903,Hu2023multilingual,min2022deep,Hu2024}.

CTC is prone to overfit the sequence order in CSLR systems, which can result in the spatial feature extraction component being under-trained ~\cite{Min2021}. To address this issue, various approaches have been suggested to improve the framework training \cite{Min2021,Papastratis2021,Hao2021,guo2023distilling}. Min et al. \cite{Min2021} introduced Visual Alignment Constraint (VAC), which inserts an auxiliary classifier to promote more rigorous training of the spatial backbone. Further advancements utilized knowledge distillation to enable the main and auxiliary classifiers to benefit from shared weights ~\cite{Hao2021}. Some researchers improved the spatial feature extractor by training it to locate important regions in the frames using spatial attention maps \cite{hu2023self} or highlighting key areas of interest using pose keypoints \cite{zuo2022c2slr}.  Additionally, the development of the Correlation Network (CorrNet) has been a significant milestone in CSLR, as it achieved SOTA results on the Phoenix2014 and CSL datasets, while leveraging the single RGB modality. The model utilized correlation maps between successive frames for effective feature extraction. Recently, Jang et al. ~\cite{jang2023self} introduced DFConv, which processes the upper and lower regions of the input frame separately to extract features from the face and body. However, this method is likely to fail with videos having signers located within various distances from the camera.  Zuo et al. \cite{Zuo2024} developed a signer-independent framework using a signer removal method which disentangles signer-specific features to create robust signer-independent CSLR models.  Several studies focused on unseen sentence recognition ~\cite{Pu2020,Hu2024}. Cross modal augmentation was proposed for gloss-level augmentation, where the gloss labels were modified along with the corresponding video frames to create new unseen sentences. However, this approach required significant manual effort to generate the new video-label pairs. Consequently, the framework was improved by utilizing a language model to create the augmented sentences \cite{Hu2024}.

 Skeleton-based CSLR was explored in ~\cite{Wang2021,Jiao_2023_ICCV,signbertplus}, leveraging key points representing the joints and bones to provide a simplified yet informative representation of the signer's movements. Skeleton data from the hands, face, and body of the signer were extracted and modeled using Spatial-Temporal GCNs (ST-GCN)  in \cite{Wang2021, Jiao_2023_ICCV}. Several studies integrated additional data sources or modalities to improve the accuracy and robustness of CSLR \cite{Zhou2021SignBERT,chen2023twostream,wei2023improving,signbertplus}. SignBERT \cite{Zhou2021SignBERT} leveraged pose data to create cropped hand images and were encoded along with the full frame images. A dynamic weighting technique was proposed to enhance the framework, resulting in significant performance improvements \cite{Zhou2022}. The study \cite{signbertplus} showed that fusing RGB and pose features significantly enhanced the performance of the network. Chen et al. ~\cite{chen2023twostream} developed a two-stream framework with 3DCNN backbones, that processes RGB frames and pose heatmap images in two streams, achieving SOTA results. The framework, however, is complex and significantly hard to train. The two-stream framework was leveraged in \cite{wei2023improving} to build a cross-lingual CSLR framework that leverages similar signs from different sign languages to enhance training with additional data. Although the method provided significant performance gains, the process of creating the extra data is tedious, requiring the creation of sign language dictionaries and isolated sign language classifiers. 

\section{CSLR Methods}\label{sec:CS}
\noindent
We empirically assess five top-performing and most commonly utilized CSLR approaches to better understand their performance on different sign languages. These models are VAC~\cite{Min2021}, Self-Mutual Distillation Learning (SMDL)~\cite{Hao2021}, Temporal Lift Pooling (TLP)~\cite{hu2022temporal}, Self-Emphasizing Network (SEN)~\cite{hu2023self}, and Correlation Network (CorrNet)~\cite{Hu2023}. 
The chosen models demonstrate various methods to extract informative features, as well as different techniques and strategies to train the feature extractor effectively without over-fitting. Moreover, the selected models were assessed on various datasets to gauge their effectiveness in modeling other sign languages. More specifically, three datasets representing three different sign languages have been chosen for the comparative evaluation, Phoenix2014~\cite{Koller2015}, ArabSign~\cite{luqman2022arabsign}, and GrSL~\cite{Adaloglou2021}.

\vspace{2mm}\noindent\textbf{Visual Alignment Constraint (VAC).} 
 Min et al. ~\cite{Min2021} noted that CTC-based CSLR models often suffer from overfitting due to inadequate training of the features extractor model. Therefore, a VAC module was introduced to enhance the feature extraction by providing alignment supervision. The features extractor module comprised 2D CNNs followed by TConvs. A pretrained ResNet-18 was used as the backbone for the CNN model. The alignment module includes Bidirectional Long Short-Term Memory (BLSTM) and a classifier trained using CTC loss with two additional losses, the visual enhancement (VE) and the visual alignment (VA) losses. The VE loss enforces the alignment between the spatial features and the target sequence, while the VA loss provides contextual supervision by focusing on long-term context predictions using knowledge distillation.

\vspace{2mm}\noindent\textbf{Self-Mutual Distillation Learning (SMDL).} 
Hao et al. ~\cite{Hao2021} claimed that the sequential learning module in CSLR frameworks tends to overfit the sequential information, such as the order of the signs, and neglects the visual features. Hence, an SMKD technique was introduced to address these issues. The method involves training the visual and sequential modules simultaneously with CTC loss by allowing them to share the same weights. The proposed method was incorporated into the VAC model ~\cite{Min2021}, where the visual module is composed of 2DCNN-TConv, and the sequential module is based on BLSTM. Additionally, this method addressed the CTC spike phenomenon, which causes the model to focus only on a few significant frames and consequently limits the effectiveness of training the feature extractor. To address this problem, the study introduced a gloss segmentation method that enhances the generalization of the model by extracting more visual information.

\vspace{2mm}\noindent\textbf{Temporal Lift Pooling. }  
Hu et al. ~\cite{hu2022temporal} argued that common pooling methods in neural networks often fail to preserve some important features when creating down-sampled feature maps. Consequently, the study proposed a TLP technique based on the Lifting Scheme. TLP involves three phases: lifting, weighting, and fusion. The lifting process produces a compressed feature representation, which is then weighted to signify the importance of each component. Then, the information is fused using a simple sum operation to generate a downsized feature map. To evaluate the proposed TLP method, the max pooling layer in the VAC model from ~\cite{Min2021} was replaced by TLP. 

\vspace{2mm}\noindent\textbf{Self-Emphasizing Network} 
The SEN was proposed in~\cite{hu2023self} to localize informative spatial regions and identify relevant frames. It is comprised of two modules: temporal self-emphasizing module (TSEM) and the spatial self-emphasizing module (SSEM). The SSEN module is designed to find the discriminative spatial features in the frame using convolutions with increasing dilations. Conversely, the TSEM module identifies which frames are important and which are not. This is achieved by finding the difference between two adjacent frames and concatenating the appearance features and the motion features. The SEN network was added to each block in the feature extractor from the VAC model in~\cite{Min2021}. The proposed modules were proven to add minimal additional computation.

\vspace{2mm}\noindent\textbf{Correlation Network (CorrNet).} 
CorrNet~\cite{Hu2023} depends on computing correlation maps of body trajectories between successive frames to identify the hand and face regions in the frames. This technique is comprised of two modules, correlation, and identification modules. The correlation module computes the correlation maps and the identification module leverages the correlation maps to dynamically emphasize or suppress the informative or noisy features. The CorrNet was implemented within the feature extractor in the 2DCNN-TConv-BLSTM model.

\section{Experiments and Results}\label{Sec:Ex}

\begin{figure} [!ht]
    \centering
    \includegraphics[width=\linewidth]{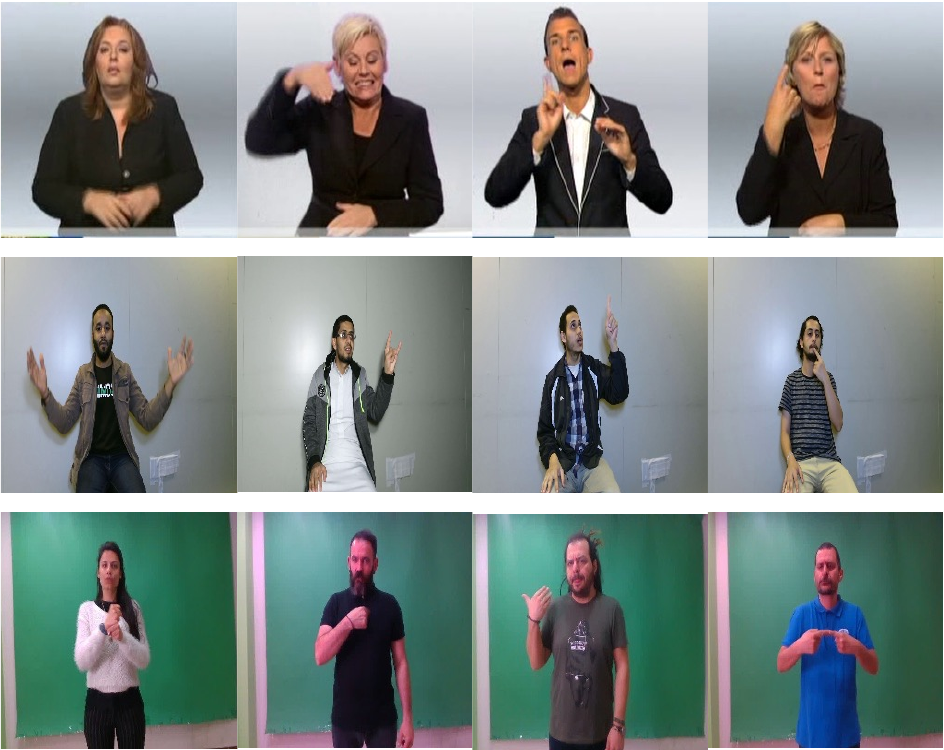}
    \caption{Samples from the utilized datasets Phoenix2014 (top), ArabSign (middle), and GrSL (bottom).}
    \label{fig:samples_datasets}
\end{figure}

\vspace{2mm}\noindent\textbf{Datasets.}
The selected models are trained and tested on three publicly available CSLR datasets Phoenix2014~\cite{Koller2015}, ArabSign~\cite{luqman2022arabsign}, and GrSL~\cite{Adaloglou2021}. Figure \ref{fig:samples_datasets} shows samples from the three utilized datasets. The Phoenix2014 dataset consists of 6,841 sentences in GSL gathered from weather forecasts. The dataset has a vocabulary size of 1,389 glosses and includes nine signers. The authors provided two evaluation sets, the multi-signer set for Signer-Dep evaluation and the Signer-Indep dataset where one signer (signer-5) is left out of the training set for testing. The ArabSign dataset contains of 9,335 samples representing 50 sentences of ArSL. Each sentence was repeated several times by six signers, and recorded in a controlled environment using a Kinect V2 camera. The continuous GrSL dataset is comprised of 331 sentences in GrSL with a vocabulary size of 310 signs. The sentences were performed by seven signers where each signer repeated each sentence five times. The dataset was recorded in a controlled environment using a RealSense camera capturing RGB and depth data. The authors provided two evaluation sets for Signer-Indep and Unseen-Sent evaluations. In the Signer-Indep split, one signer is left out for the testing, namely signer-3, and the remaining signers are used for training. In the Unseen-Sent set, 10\% of the sentences are excluded from the training data and reserved for testing.

\vspace{2mm}\noindent\textbf{Experiments Setup.}
We followed the same experimental settings described in each study to compare the models fairly. The experiments were conducted on an NVIDIA RTX A6000 with two 48GB GPUs. The models were trained using Adam optimization for 40 epochs with a batch size of 2 and an initial learning rate of $10^{-3}$, which was reduced at epochs 20 and 35 by a factor of five. For data augmentation, the video frames were resized to 256x256 and randomly cropped to 224X224. Additionally, random temporal scaling was applied to 20\% of the frames, and horizontal flipping was performed with 50\% probability. Word Error Rate (WER) was utilized to evaluate the models, which computes the number of edits needed to transfer the predicted sentence into the actual ground truth sentence. The WER is calculated using the following formula: 
\begin{equation} \label{eq:wer}
WER = \frac{S + D + I}{N} \end{equation}
where $S$ is the number of substitutions, $D$ is the number of deletions, $I$ is the number of insertions, and N is the number of words in the reference sentence (the correct sequence of words). The WER returns a value between 0 and 1, and the lower the WER, the better the performance of the CSLR model.

\begin{figure*} []
    \centering
    \begin{subfigure}[t]{0.3\textwidth}
        \centering
        \includegraphics[width=\textwidth]{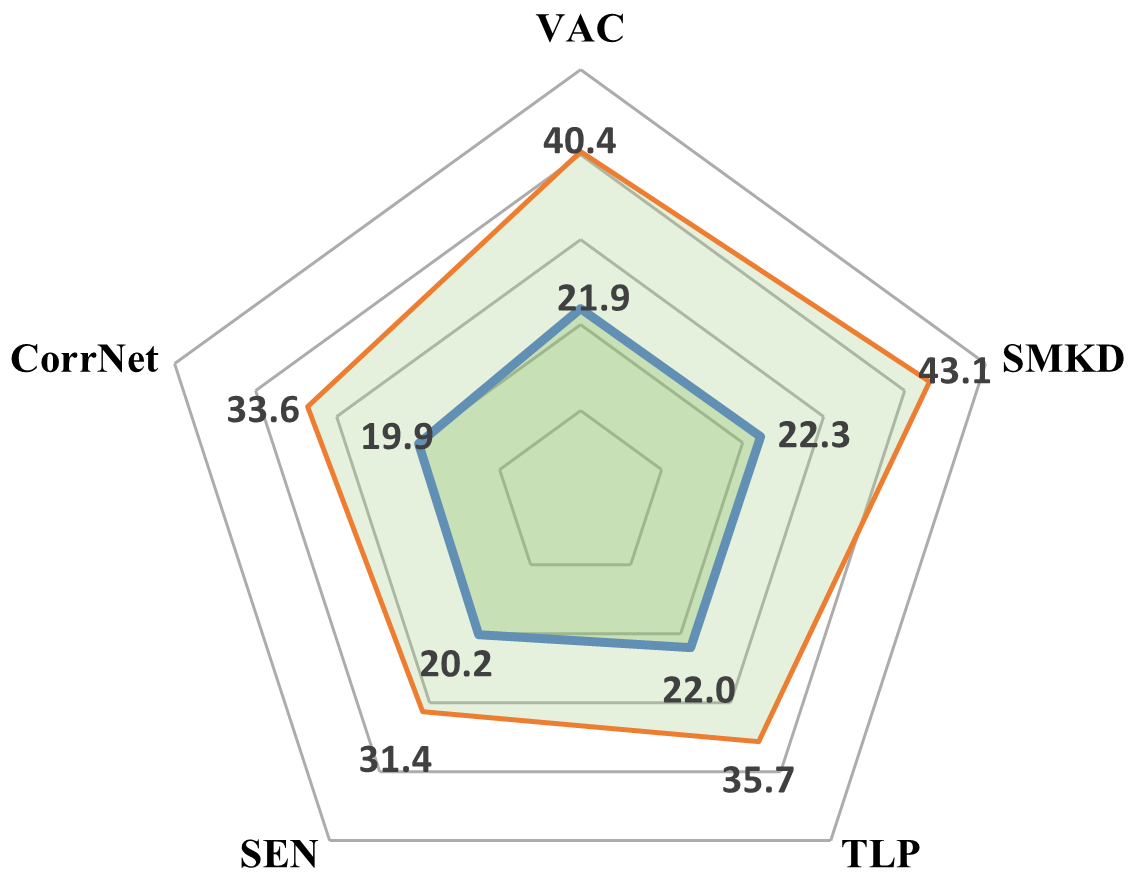}
        \caption{}
    \end{subfigure}%
   \hspace{1em}  
    \begin{subfigure}[t]{0.3\textwidth}
        \centering
        \includegraphics[width=\textwidth]{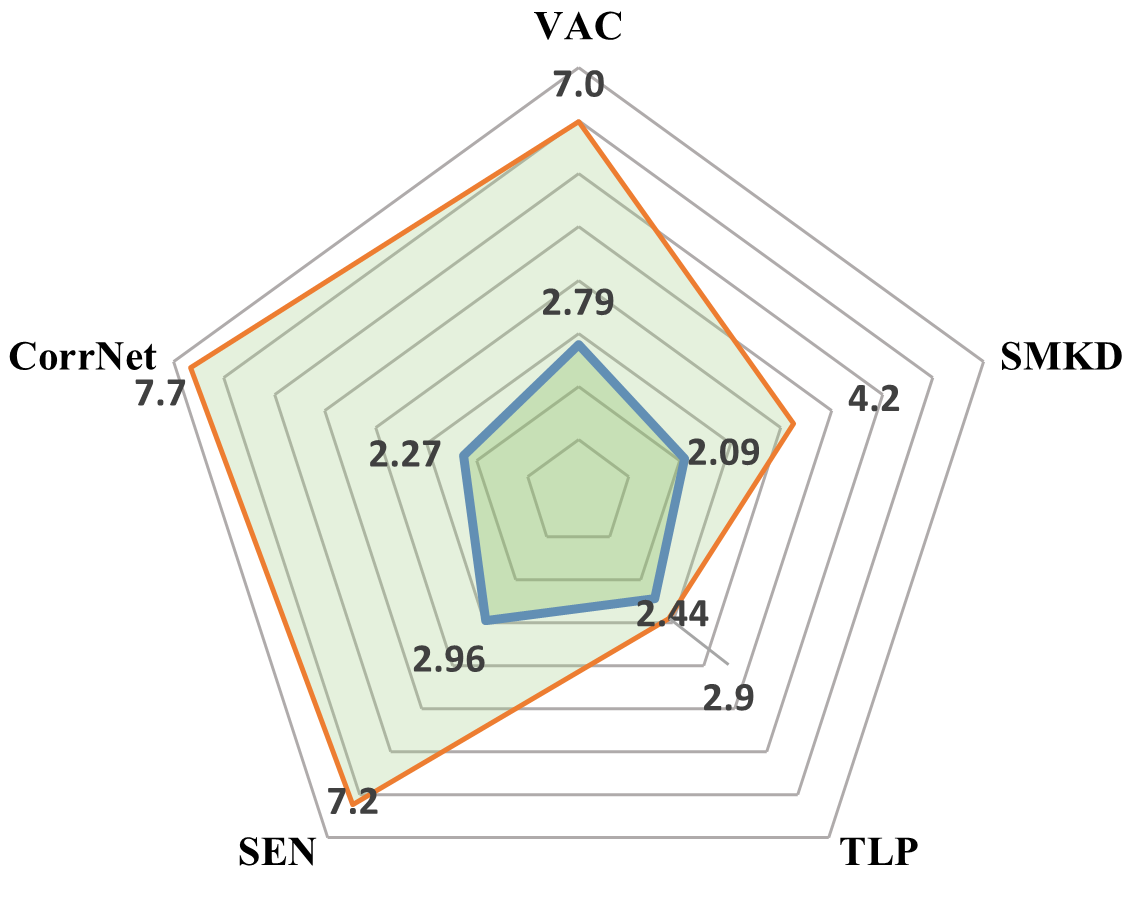}
        \caption{}
    \end{subfigure}%
    \hspace{1em}
        \begin{subfigure}[t]{0.3\textwidth}
        \centering
        \includegraphics[width=\textwidth]{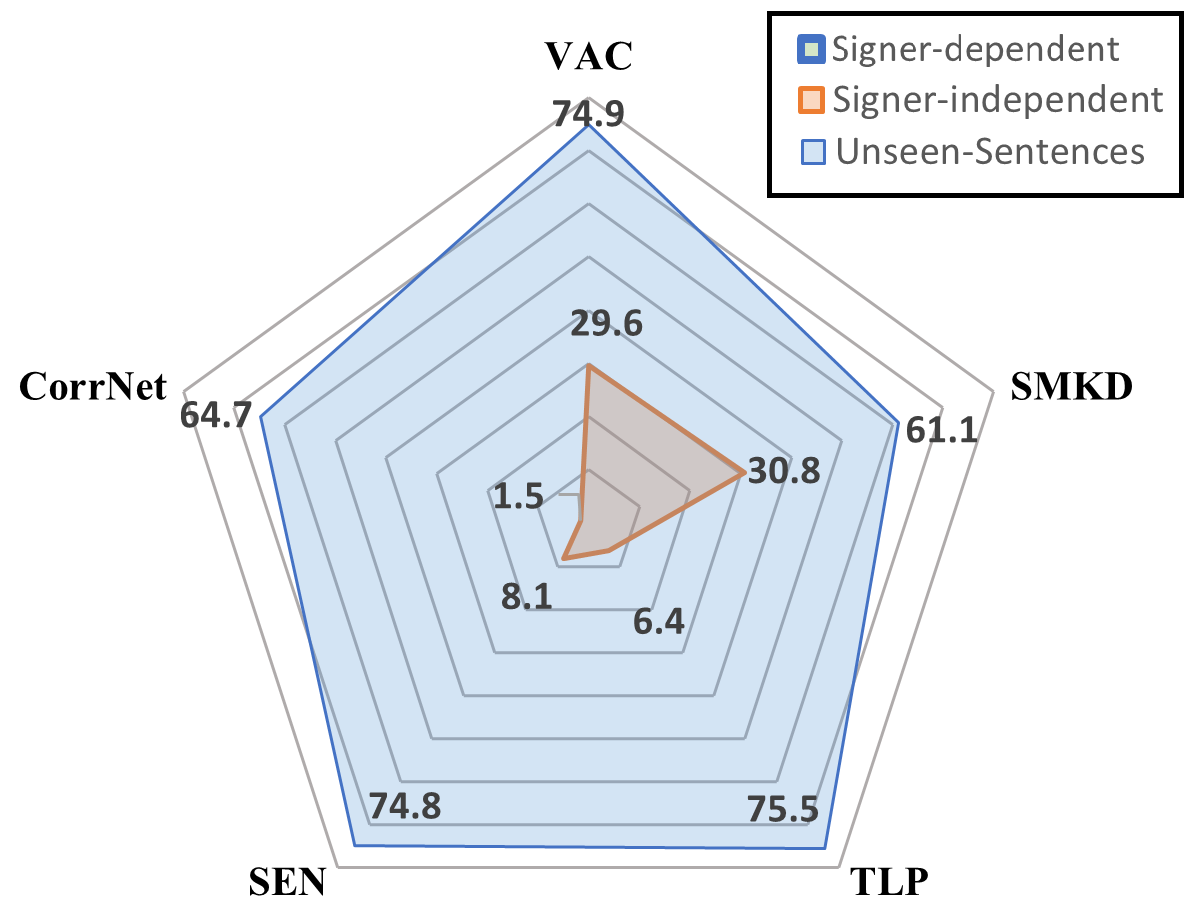}
        \caption{}
    \end{subfigure}
    \caption{WERs of the evaluated models in different evaluation settings using (a) Phoenix2014, (b) ArabSign, and (c) GrSL datasets.}
        \label{fig:results_chart}

\end{figure*}

\vspace{2mm}\noindent\textbf{Results Analysis and Discussion.}
The WERs of the evaluated techniques in the Signer-Dep, Signer-Indep,  and Unseen-Sent evaluation settings are shown in Table \ref{tab:results} and Figure \ref{fig:results_chart}.  For the Phoenix2014 dataset in the Signer-Dep mode, we reproduced the results for the VAC, TLP, SMKD, SEN, and CorrNet models as reported by the authors. We achieved similar results with around 1\% fluctuations in WERs. The CorrNet model achieved the best performance, which is in line with the results reported by the CorrNet authors ~\cite{Hu2023}. This is primarily due to the extra attention put into identifying the hand and face regions through the correlation maps. The SEN is the second best performing model, obtaining 20.2\% WER. We noticed in our experiments that the SEN model obtained 1\% better WER than the results reported by the authors in ~\cite{hu2023self}. Third is the VAC model with 21.9\% WER. The TLP and SMKD models delivered similar performance, around 22\% WER. 

We also report the Signer-Indep evaluation's results for the Phoenix2014 dataset using the Signer-Indep dataset split, which was not reported by the authors of the five models. Signer-Indep evaluation is more challenging than Signer-Dep as revealed by the increased WERs. In this split, the CorrNet model also provided the best results, followed by SEN, TLP, VAC, and SMKD. Notably, SMKD provided the worst performance on the Phoenix2014 dataset in both Signer-Dep and Signer-Indep evaluations. 

For the ArabSign dataset, the performance of the evaluated models in the Signer-Dep mode is very similar with around 0.2\% WER. This is because the ArabSign dataset is relatively small, with only 50 sentences which was not a challenging task for these models. As for Signer-Indep evaluation, since the ArabSign dataset does not have a specific set for Signer-Indep evaluation, we follow the same setup used in ~\cite{luqman2022arabsign}, where we leave out one signer for testing in each trial. This is repeated six times since the dataset contains six signers. We also report the average WER amongst all signers. The Signer-Indep results for the ArabSign dataset are reported in Table \ref{tab:SI_ArabSign}, which reveals that the TLP model outperformed the rest of the models with 2.9\% WER. Surprisingly, CorrNet delivered the worst performance with this set, which suggests that the model was likely overfitting the small training data and struggled to generalize with the unseen signer during evaluation. Better results are likely to be obtained by fine-tuning the training parameters using a smaller learning rate and adopting early stopping.

On the other hand, the recognition of new sentences (Unseen-Sent) not seen in the training was proven to be a difficult task, as portrayed by the obtained high WERs on the GrSL Unseen-Sent split. The SKMD model outperformed the other models with 61.1\% WER, followed by the CorrNet model with 64.7\% WER. This indicates that knowledge sharing between the spatial and temporal modules in SMKD strongly assisted recognizing new sign sequences in this task. Regarding the GrSL Signer-Indep assessment, CorrNet excelled with the Signer-Indep scenario with a substantially lower WER of 1.5\%, which is far ahead of the other models. Overall, CorrNet stands out in most settings across the datasets, indicating its high robustness and generalizability. TLP also shows consistent performance across Signer-Indep settings, which implies it is generally more adaptable to Signer-Indep scenarios. This signifies the effectiveness of the TLP layer in creating more representative features. SMKD, however, performs well with new sentences and Signer-Dep settings, which may indicate specialization in scenarios where the signing style is consistent. 
 
\begin{table*}[t]
\centering
\footnotesize
\caption{WERs of the evaluated models in different settings.}
\label{tab:results}
\begin{tabular}{lcccccc}
\toprule
\multirow{2}{*}{Model}                     & \multicolumn{2}{c}{Phoenix2014} & \multicolumn{2}{c}{ArabSign} & \multicolumn{2}{c}{GrSL} \\ \cmidrule{2-7} 
                                    & Signer-Dep       & Signer-Indep     & Signer-Dep      & Signer-Indep            & Signer-Indep & Unseen-Sent    \\ \midrule
VAC     & 21.94 & 40.40 & 0.2787 & 6.97 & 29.60 & 74.92  \\
SMKD    & 22.31 & 43.08 & \textbf{0.2091} & 4.24  & 30.80 & \textbf{61.10} \\
TLP     & 22.00 & 35.70 & 0.2439 & \textbf{2.90}   & 6.40 & 75.50 \\
SEN     & 20.20 & 34.40 &  0.2962 & 7.22  & 8.09 & 74.80  \\
CorrNet & \textbf{19.90}    & \textbf{33.60} & 0.2265 & 7.65     & \textbf{1.50} & 64.70 \\

\bottomrule
\end{tabular}
\end{table*}

\begin{table}[]
\centering
\footnotesize
\caption{WERs of the evaluated models in Signer-Indep mode using the ArabSign dataset.}

\label{tab:SI_ArabSign}
\begin{tabular}{@{}lccccccc@{}}
\toprule
\multirow{2}{*}{Model} & \multicolumn{6}{c}{Tested Signer}                    &        \\ \cmidrule(l){2-8} 

               & 1       & 2       & 3       & 4       & 5       & 6       & Avg    \\ \midrule
VAC                 & 7.55  & 5.03  & 8.67  & 5.66  & 8.17  & 6.76  & 6.97  \\
SMKD                & 5.88  & 2.93  & 2.11  & 4.52  & 1.89  & 8.11  & 4.24  \\
TLP                 & 2.91  & 0.53  & 2.09  & 1.14  & 1.29  & 9.46  & \textbf{2.90}  \\
SEN                 & 5.84  & 5.70  & 6.20  & 10.90 & 5.40  & 9.30  & 7.22  \\
CorrNet             & 10.50  & 5.50  & 6.70  & 11.10 & 3.80  & 8.30  & 7.65  \\
 \bottomrule
\end{tabular}
\end{table}

\begin{figure*} [t]
    \centering
    \includegraphics[width=\linewidth]{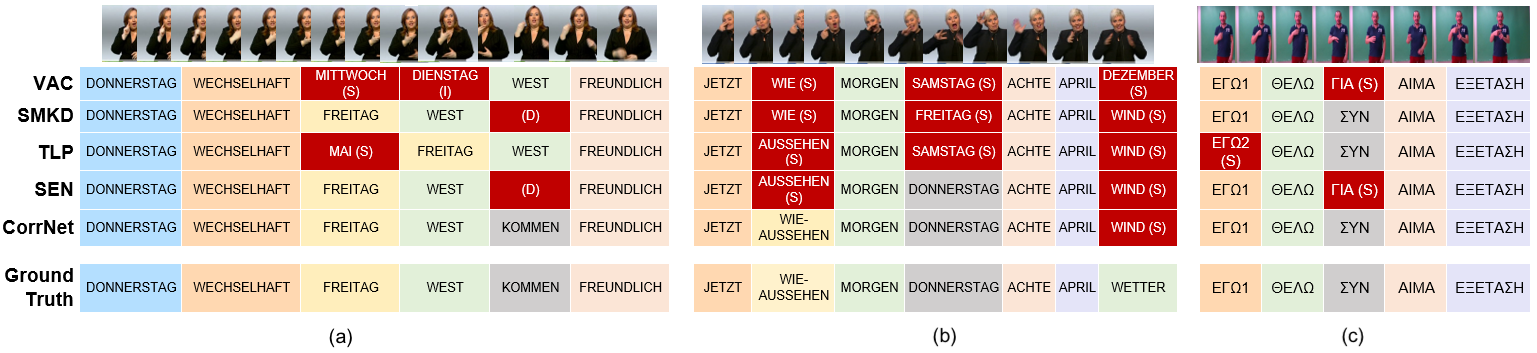}
    \caption{Samples of gloss predictions of evaluated models with (a) Signer-Dep, (b) Signer-Indep, and (c) Unseen-Sent evaluation settings. Errors are colored in pink, where (D) indicates a deletion error, (I) an insertion error, and (S) a substitution error.}
    \label{fig:errors}
\end{figure*}

\begin{figure} [t]
    \centering
    \includegraphics[width=\linewidth]{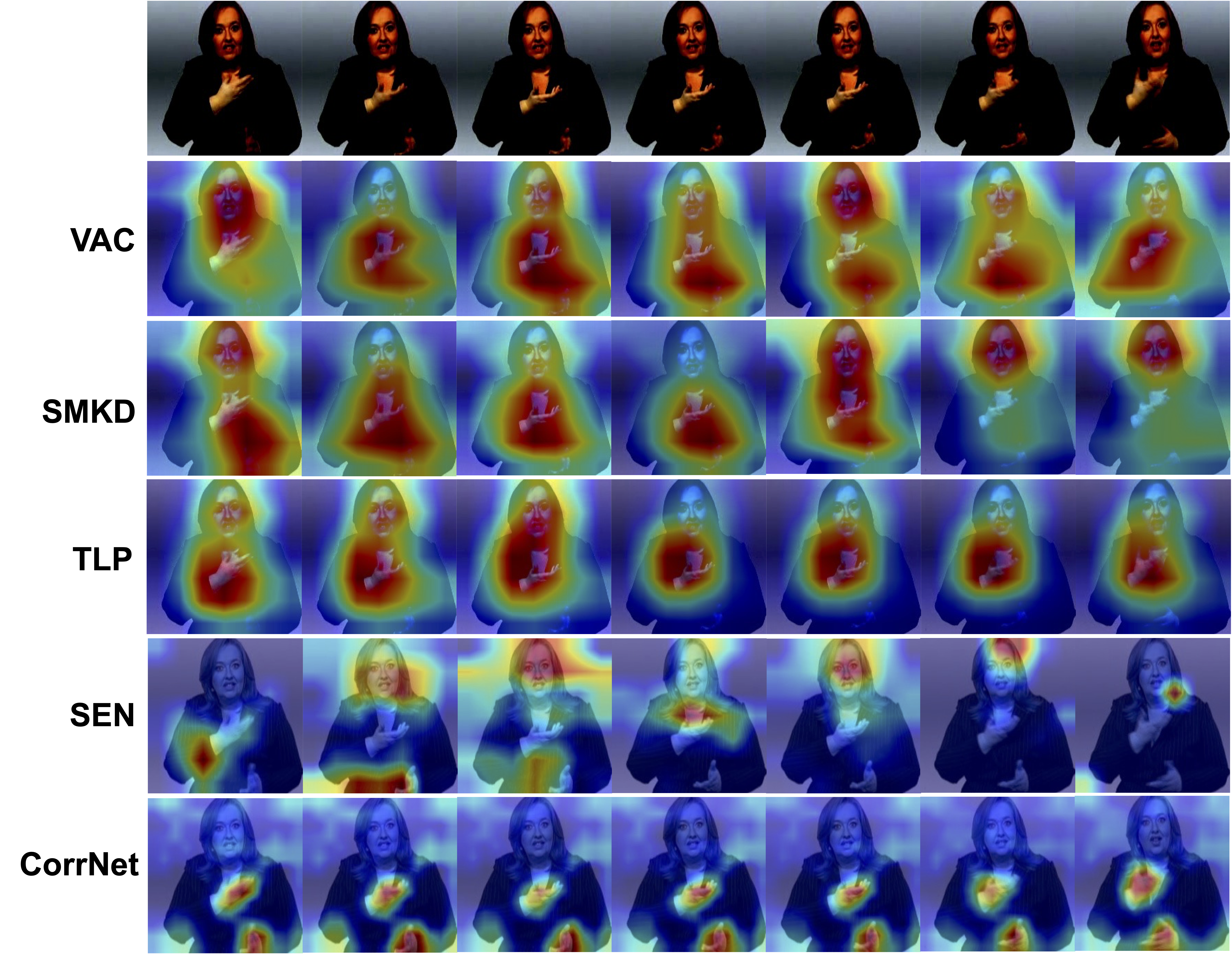}
    \caption{GradCam \cite{selvaraju2017grad} visualization of gloss \textit{"KOMMEN"} by the evaluated models on Phoenix2014 Signer-Dep. The red areas indicate the most attended to regions.  }
    \label{fig:gradcam}
\end{figure}

To further understand the recognition ability of the evaluated models, we show in Figure \ref{fig:errors} the predicted glosses by the evaluated models in different evaluation settings. In the Signer-Dep setting, CorrNet, unlike the other models, was able to correctly recognize all the sentence's glosses, whereas all the models failed to recognize the \textit{KOMMEN} (come) gloss. To investigate this discrepancy, we generated GradCam visualizations \cite{selvaraju2017grad} corresponding to the \textit{KOMMEN} sign by the evaluated models as shown in Figure \ref{fig:gradcam}. In the analyzed sample, the signer used both the right and left hand to portray the sign. However, only CorrNet was able to focus on both hands even though the left hand movements were subtle and the hand was partially out of the frame. This provides a potential explanation for its success in recognizing the challenging sign. 

\begin{figure} [t]
    \centering
    \includegraphics[width=.7\linewidth]{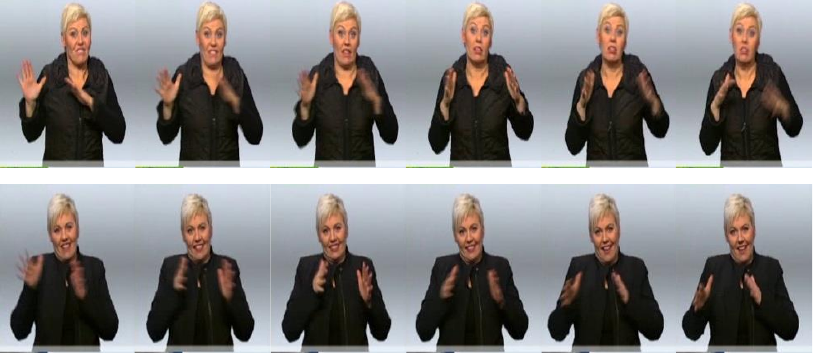}
    \caption{Demonstrating similar signs \textit{"WIND"} (top row) and \textit{"WETTER"} (bottom row) performed by signer-5 from the Phoenix2014 dataset.}
    \label{fig:WindVsWetter}
\end{figure}

\begin{figure} [t]
    \centering
    \includegraphics[width=.6\linewidth]{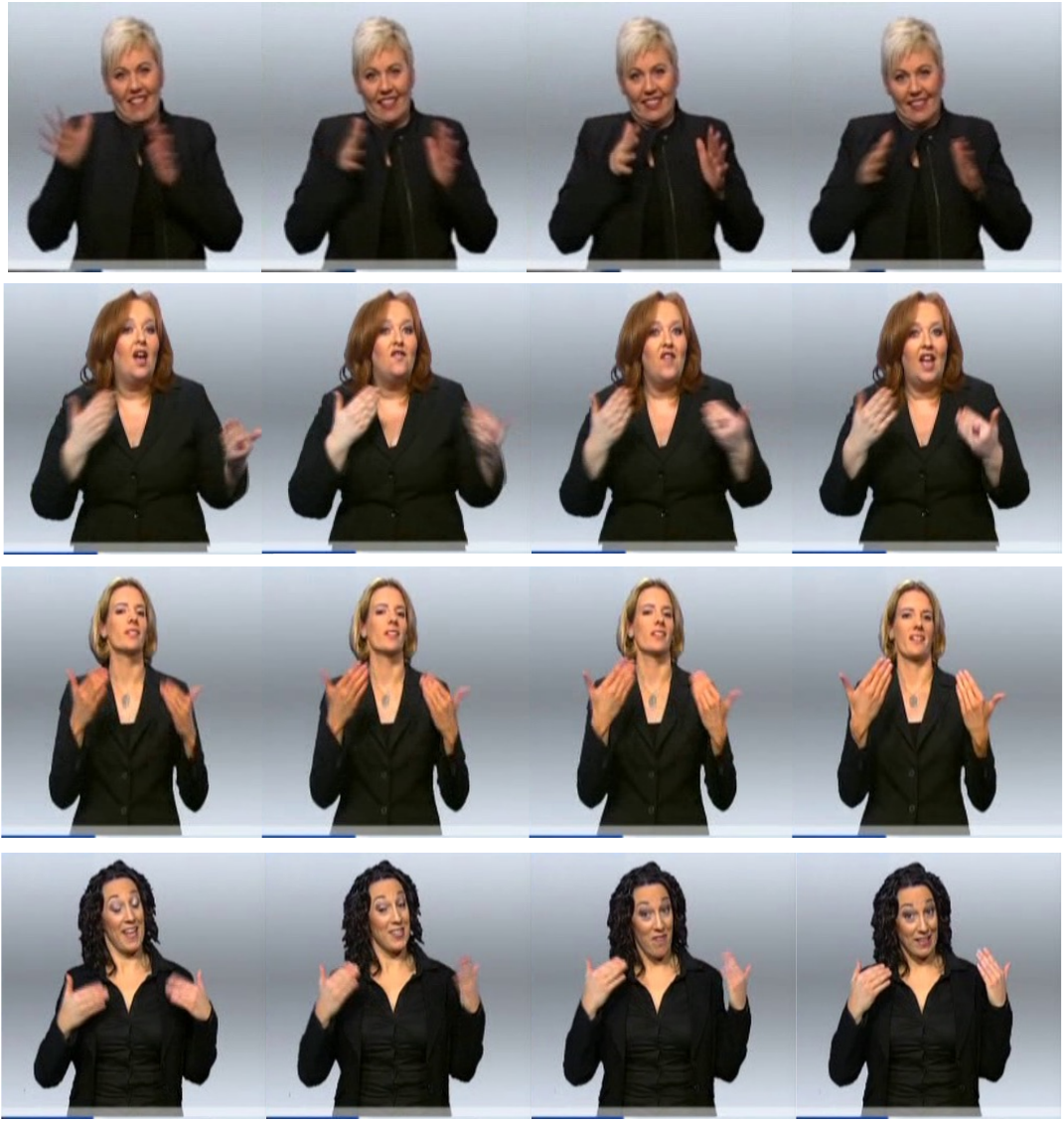}
    \caption{Showcasing different styles of sign \textit{"WETTER"} by signer-5 (top row) and other signers (rows 2,3 and 4) from the Phoenix2014 dataset.}
    \label{fig:Wetter}
\end{figure}

CorrNet also outperformed the rest of the models with the Signer-Indep evaluation. However, we can see that all the models, including CorrNet, were not able to identify the \textit{WETTER} (Weather) sign (Figure \ref{fig:gradcam} (b)), as most of the models mistook it with the sign \textit{WIND}. This misclassification can be attributed to the inherent similarity between the two signs, as illustrated in Figure \ref{fig:WindVsWetter}. Further investigation revealed that the signer evaluated in the Phoenix2014 Signer-Indep dataset (signer-5) performed the \textit{"WETTER"} sign differently from other signers, with palms directed inward toward the face, as depicted in Figure \ref{fig:Wetter}. This highlights the significant challenge in developing Signer-Indep CSLR systems capable of generalizing across various signing styles. 
For Unseen-Sent prediction, CorrNet and SMKD both correctly predicted the new sign sequence, affirming their superior performance in interpreting sign language sentences which they have not been previously exposed to. In summary, the qualitative analysis further validates the superiority of the CorrNet model with different testing scenarios and across a variety of datasets and sign languages. The results also emphasize that building robust models that can generalize with both unseen signers and unseen sentences is a difficult task.


\section{Conclusion} \label{sec:conc}
\noindent
To provide a deep understanding on the performance of video based CSLR methods, this paper presented a comparative evaluation of five recently proposed deep learning CSLR models, namely VAC, SMKD, TLP, SEN and CorrNet. The models were evaluated using various challenging settings, including unseen signers and sentences. Further, to gauge their performance on different sign languages, the models were assessed using three datasets, Phoenix2014 , ArabSign and GrSL. As a results, we established new benchmarks on the targeted datasets and analyzed their performance accordingly. The findings of this study showed that there are inherent trade-offs between adapting to signer variability and handling unseen sentences. Nevertheless, CorrNet was the most robust model across different settings and sign languages, due to its supreme features modeling capability. Future research can include a broader category of SLR frameworks including finger-spelling, isolated and sign language translation models. 



\section*{Acknowledgments}
\noindent The authors would like to acknowledge the support received from the Saudi Data and AI Authority (SDAIA) and King Fahd University of Petroleum and Minerals (KFUPM) under the SDAIA-KFUPM Joint Research Center for Artificial Intelligence Grant JRC-AI-RFP-14.



\end{document}